\documentclass[11pt,letterpaper]{article}
\usepackage{acl2015}
\usepackage{times}
\usepackage{latexsym}
\usepackage[utf8]{inputenc}
\usepackage{amsmath}
\usepackage{amsthm}
\usepackage{amssymb}
\usepackage{graphicx}
\usepackage{color}
\usepackage{hhline}
\usepackage{url}
\usepackage{multirow}

\usepackage[normalem]{ulem}

\newcommand{\bmx}[0]{\begin{bmatrix}}
\newcommand{\emx}[0]{\end{bmatrix}}

\newcommand{\vect}[1]{\mathbf{#1}}
\newcommand{\vects}[1]{\boldsymbol{#1}}

\newcommand{\vw}[0]{\vect{w}}

\newcommand{\TT}[0]{\vects{\theta}}

\newcommand{\E}[0]{\mathbb{E}}

\DeclareMathOperator*{\argmax}{\arg \max}

\newcommand{\ola}{\overleftarrow}
\newcommand{\ora}{\overrightarrow}

\newcommand{\bracket}[1]{\ensuremath{\left<#1\right>}}

\graphicspath{ {./figures/} }

\title{On Using Very Large Target Vocabulary for Neural Machine Translation}

\author{S\'ebastien Jean\\
        Universit\'e de Montr\'eal 
        \And
        Kyunghyun Cho \\
        Universit\'e de Montr\'eal 
        \And 
        Roland Memisevic\\
	    Universit\'e de Montr\'eal
	  \And
	Yoshua Bengio\\
	    Universit\'e de Montr\'eal \\
            CIFAR Senior Fellow
        }

\date{}

\begin{document}

\maketitle

\begin{abstract}
    Neural machine translation, a recently proposed approach to machine
    translation based purely on neural networks, has shown promising results
    compared to the existing approaches such as phrase-based statistical
    machine translation. Despite its recent success, neural machine translation
    has its limitation in handling a larger vocabulary, as training complexity
    as well as decoding complexity increase proportionally to the number of
    target words. In this paper, we propose a method based on importance
    sampling that allows us to use a very large target vocabulary without
    increasing training complexity.  We show that decoding can be efficiently
    done even with the model having a very large target vocabulary by selecting
    only a small subset of the whole target vocabulary.
    The models trained by the proposed approach
    are empirically found to match, and in some cases outperform, the baseline models
    with a small vocabulary as well as the LSTM-based neural machine translation  
    models.  Furthermore, when we use an
    ensemble of a few models with very large target vocabularies, we achieve     
    performance comparable to the state of the art (measured by BLEU) on both
    the English$\to$German and English$\to$French translation tasks of WMT'14.
\end{abstract}

\section{Introduction}

Neural machine translation (NMT) is a recently introduced approach to solving
machine
translation~\cite{Kalchbrenner2013,Bahdanau-et-al-arxiv2014,Sutskever-et-al-NIPS2014}.
In neural machine translation, one builds a single neural network that reads
a source sentence and generates its translation. The whole neural network is
jointly trained to maximize the conditional probability of a correct translation
given a source sentence, using the bilingual corpus. The NMT models have shown
to perform as well as the most widely used conventional translation
systems~\cite{Sutskever-et-al-NIPS2014,Bahdanau-et-al-arxiv2014}.

Neural machine translation has a number of advantages over the existing
statistical machine translation system, specifically, the phrase-based
system~\cite{Koehn2003}. First, NMT requires a minimal set of domain knowledge.
For instance, all of the models proposed in \cite{Sutskever-et-al-NIPS2014},
\cite{Bahdanau-et-al-arxiv2014} or \cite{Kalchbrenner2013} do not assume any
linguistic property in both source and target sentences except that they are
sequences of words. Second, the whole system is jointly tuned to maximize the
translation performance, unlike the existing phrase-based system which consists
of many feature functions that are tuned separately. Lastly, the memory
footprint of the NMT model is often much smaller than the existing system which
relies on maintaining large tables of phrase pairs.

Despite these advantages and promising results, there is a major
limitation in NMT compared to the existing phrase-based approach. That is,
the number of target words must be limited. This is mainly because the
complexity of training and using an NMT model increases as the number of target
words increases. 

A usual practice is to construct a target vocabulary of the $k$ most frequent words
(a so-called shortlist), where $k$ is often in the range of
$30,000$~\cite{Bahdanau-et-al-arxiv2014} to
$80,000$~\cite{Sutskever-et-al-NIPS2014}. Any word not included in this
vocabulary is mapped to a special token representing an {\it unknown} word
$\left[ \mbox{UNK} \right]$. This approach works well when there are only a few
unknown words in the target sentence, but it has been observed that the
translation performance degrades rapidly as the number of unknown words
increases~\cite{Cho2014a,Bahdanau-et-al-arxiv2014}.

In this paper, we propose an approximate training algorithm based on (biased)
importance sampling that allows us to train an NMT model with a much larger
target vocabulary. The proposed algorithm effectively keeps the computational
complexity during training at the level of using only a small subset of the full
vocabulary. Once the model with a very large target vocabulary is trained, one
can choose to use either all the target words or only a subset of them.

We compare the proposed algorithm against the baseline shortlist-based approach
in the tasks of English$\to$French and English$\to$German translation using the
NMT model introduced in \cite{Bahdanau-et-al-arxiv2014}. The empirical results
demonstrate that we can potentially achieve better translation performance using
larger vocabularies, and that our approach does not sacrifice too much speed for both
training and decoding.  Furthermore, we show that the model trained with this
algorithm gets the best translation performance yet achieved by single NMT
models on the WMT'14 English$\to$French translation task.

\section{Neural Machine Translation and Limited Vocabulary Problem}

In this section, we briefly describe an approach to neural machine translation
proposed recently in \cite{Bahdanau-et-al-arxiv2014}. Based on this description
we explain the issue of limited vocabularies in neural machine translation.

\subsection{Neural Machine Translation}

Neural machine translation is a recently proposed approach to machine
translation, which uses a single neural network trained jointly to maximize the
translation performance
\cite{Forcada1997,Kalchbrenner2013,Cho-et-al-EMNLP2014,Sutskever-et-al-NIPS2014,Bahdanau-et-al-arxiv2014}.

Neural machine translation is often implemented as the encoder--decoder network.
The encoder reads the source sentence $x=(x_1, \dots, x_{T})$ and encodes it
into a sequence of hidden states $h=(h_1, \cdots, h_{T})$:
\begin{align}
    \label{eq:generic_encoder}
    h_t = f\left( x_{t}, h_{t-1} \right).
\end{align}
Then, the decoder, another recurrent neural network,  generates a corresponding
translation $y=(y_1, \cdots, y_{T'})$ based on the encoded sequence of hidden
states $h$:
\begin{align}
    \label{eq:generic_output}
    p(y_t \mid y_{<t}, x) \propto \exp\left\{ q\left( y_{t-1},
    z_{t}, c_t\right) \right\},
\end{align}
where 
\begin{align}
    \label{eq:generic_decoder}
    z_t = g\left( y_{t-1}, z_{t-1}, c_t\right), \\
    \label{eq:generic_context}
    c_t = r\left( z_{t-1}, h_1, \dots, h_{T} \right),
\end{align}
and $y_{<t} = (y_1, \dots, y_{t-1})$.

The whole model is jointly trained to maximize the conditional log-probability of
the correct translation given a source sentence with respect to the parameters
$\TT$ of the model:
\begin{align*}
    \TT^* = \argmax_{\TT} \sum_{n=1}^N \sum_{t=1}^{T_n} \log p(y_t^n \mid y_{<t}^n, x^n),
\end{align*}
where $(x^n, y^n)$ is the $n$-th training pair of sentences, and $T_n$ is the
length of the $n$-th target sentence ($y^n$).

\subsubsection{Detailed Description}
\label{sec:rnnsearch}

In this paper, we use a specific implementation of neural machine translation
that uses an attention mechanism, as recently proposed in
\cite{Bahdanau-et-al-arxiv2014}. 

In \cite{Bahdanau-et-al-arxiv2014}, the encoder in
Eq.~\eqref{eq:generic_encoder} is implemented by a bi-directional recurrent
neural network such that 
\begin{align*}
    h_t = \left[ \ola{h}_t ; \ora{h}_t\right],
\end{align*}
where
\begin{align*}
    \ola{h}_t = f\left( x_{t}, \ola{h}_{t+1}\right), \ora{h}_t = f\left( x_{t}, \ora{h}_{t-1}\right).
\end{align*}
They used a gated recurrent unit for $f$ (see, e.g., \cite{Cho-et-al-EMNLP2014}).

The decoder, at each time, computes the context vector $c_t$ as a convex sum of
the hidden states $(h_1, \dots, h_T)$ with the coefficients $\alpha_1, \dots,
\alpha_T$ computed by
\begin{align}
    \label{eq:alignment_weight}
    \alpha_t = \frac{\exp \left\{ a\left( h_t, z_{t-1} \right)\right\}}{
    \sum_k \exp \left\{ a\left( h_k, z_{t-1} \right)\right\}},
\end{align}
where $a$ is a feedforward neural network with a single hidden layer.

A new hidden state $z_t$ of the decoder in Eq.~\eqref{eq:generic_decoder} is
computed based on the previous hidden state $z_{t-1}$, previous generated symbol
$y_{t-1}$ and the computed context vector $c_t$. The decoder also uses the gated
recurrent unit, as the encoder does.

The probability of the next target word in Eq.~\eqref{eq:generic_output} is then
computed by 
\begin{align}
    \label{eq:real_output}
    p(y_t & \mid y_{<t}, x) 
    = \frac{1}{Z} \exp\left\{ \vw_t^\top \phi\left( y_{t-1},
    z_{t}, c_t\right) + b_t \right\}, 
\end{align}
where $\phi$ is an affine transformation followed by a nonlinear activation, and
$\vw_t$ and $b_t$ are respectively the {\it target word vector} and the target
word bias. $Z$ is the normalization constant computed by
\begin{align}
    \label{eq:real_output_Z}
    Z = \sum_{k:y_k \in V} \exp\left\{ \vw_k^\top \phi\left( y_{t-1},
    z_{t}, c_t\right) + b_k \right\},
\end{align}
where $V$ is the set of all the target words.

For the detailed description of the implementation, we refer the reader to the
appendix of \cite{Bahdanau-et-al-arxiv2014}.

\subsection{Limited Vocabulary Issue and Conventional Solutions}
\label{sec:related_work}

One of the main difficulties in training this neural machine translation model
is the computational complexity involved in computing the target word
probability (Eq.~\eqref{eq:real_output}). More specifically, we need to compute
the dot product between the feature $\phi\left( y_{t-1}, z_{t}, c_t\right)$ and
the word vector $w_t$ as many times as there are words in a target vocabulary in
order to compute the normalization constant (the denominator in
Eq.~\eqref{eq:real_output}). This has to be done for, on average, 20--30 words
per sentence, which easily becomes prohibitively expensive even with a
moderate number of possible target words. Furthermore, the memory requirement
grows linearly with respect to the number of target words. This has been a major
hurdle for neural machine translation, compared to the existing non-parametric
approaches such as phrase-based translation systems. 

Recently proposed neural machine translation models, hence, use a shortlist of
30,000 to 80,000 most frequent words
\cite{Bahdanau-et-al-arxiv2014,Sutskever-et-al-NIPS2014}.  This makes training
more feasible, but comes with a number of problems. First of all, the
performance of the model degrades heavily if the translation of a source
sentence requires many words that are not included in the shortlist
\cite{Cho2014a}. This also affects the performance evaluation of the system
which is often measured by BLEU. Second, the first issue becomes more
problematic with languages that have a rich set of words such as German or other
highly inflected languages.

There are two {\it model-specific} approaches to this issue of large target
vocabulary. The first approach is to stochastically approximate the target word
probability. This has been used proposed recently in
\cite{Mnih2013,Mikolov-et-al-ICLR2013} based on noise-contrastive estimation
\cite{Gutmann+Hyvarinen-2010}.  In the second approach, the target words are
clustered into multiple classes, or hierarchical classes, and the target
probability $p(y_t|y_{<t}, x)$ is factorized as a product of the class probability
$p(c_t|y_{<t}, x)$ and the intra-class word probability $p(y_t|c_t, y_{<t}, x)$. This
reduces the number of required dot-products into the sum of the number of
classes and the words in a class. These approaches mainly aim at reducing the
computational complexity during training, but do not often result in speed-up
when decoding a translation during test time\footnote{This is due to the fact
    that the beam search requires the conditional probability of {\it every}
    target word at each time step regardless of the parametrization of the
output probability.}.

Other than these model-specific approaches, there exist {\it
translation-specific} approaches. A translation-specific approach exploits
the properties of the rare target words. For instance, Luong et al. of
\cite{Luong2014} proposed such an approach for neural machine
translation. They replace rare words (the words that are not included in the
shortlist) in both source and target sentences into corresponding
\bracket{\mbox{OOV}_n} tokens using the word alignment model. Once a source
sentence is translated, each \bracket{\mbox{OOV}_n} in the
translation will be replaced based on the source word marked by the
corresponding \bracket{\mbox{OOV}_n}.

It is important to note that the model-specific approaches and the
translation-specific approaches are often complementary and can be used
together to further improve the translation performance and reduce the
computational complexity. 

\section{Approximate Learning Approach to Very Large Target Vocabulary}

\subsection{Description}
\label{sec:algorithm}

In this paper, we propose a {\it model-specific} approach that allows us to
train a neural machine translation model with a very large target vocabulary.
With the proposed approach, the computational complexity of training becomes
constant with respect to the size of the target vocabulary. Furthermore, the
proposed approach allows us to efficiently use a fast computing device with
limited memory, such as a GPU, to train a neural machine translation model with a
much larger target vocabulary.

As mentioned earlier, the computational inefficiency of training a neural
machine translation model arises from the normalization constant in
Eq.~\eqref{eq:real_output}.  In order to avoid the growing complexity of
computing the normalization constant, we propose here to use only a small subset
$V'$ of the target vocabulary at each update. The proposed approach is based on the
earlier work of \cite{Bengio+Senecal-2008}.

Let us consider the gradient of the log-probability of the output in
Eq.~\eqref{eq:real_output}. The gradient is composed of a positive and negative
part:
\begin{align}
    \label{eq:grad}
    \nabla \log p& (y_t \mid y_{<t}, x) \\
    =& \nabla \mathcal{E}(y_t) - \sum_{k: y_k
    \in V} p(y_k \mid y_{<t}, x) \nabla \mathcal{E}(y_k),
    \nonumber
\end{align}
where we define the energy $\mathcal{E}$ as
\begin{align*}
    \mathcal{E}(y_j) = \vw_j^\top \phi\left( y_{j-1}, z_{j}, c_j\right) + b_j.
\end{align*}
The second, or negative, term of the gradient is in essence the expected
gradient of the energy: 
\begin{align}
    \label{eq:neg_grad}
    \E_P \left[ \nabla \mathcal{E}(y) \right],
\end{align} 
where $P$ denotes $p(y \mid y_{<t}, x)$.  

The main idea of the proposed approach is to approximate this expectation, or
the negative term of the gradient, by importance sampling with a small number of
samples. Given a predefined proposal distribution $Q$ and a set $V'$ of samples
from $Q$, we approximate the expectation in Eq.~\eqref{eq:neg_grad} with
\begin{align}
    \label{eq:neg_grad_approx}
    \E_P \left[ \nabla \mathcal{E}(y) \right] \approx
    \sum_{k: y_k \in V'} \frac{\omega_k}{\sum_{k': y_{k'} \in V'} \omega_{k'}} \nabla
    \mathcal{E}(y_k),
\end{align}
where 
\begin{align}
    \label{eq:importance_weight}
    \omega_k = \exp \left\{ \mathcal{E}(y_k) - \log Q(y_k) \right\}.
\end{align}

This approach allows us to compute the normalization constant during training
using only a small subset of the target vocabulary, resulting in much lower
computational complexity for each parameter update. Intuitively, at each
parameter update, we update only the vectors associated with the correct word
$\vw_t$ and with the sampled words in $V'$.  Once training is over, we can
use the full target vocabulary to compute the output probability of each
target word.

Although the proposed approach naturally addresses the computational complexity,
using this approach naively does not guarantee that the number of parameters
being updated for each sentence pair, which includes multiple target words, is
bounded nor can be controlled. This becomes problematic when training is done,
for instance, on a GPU with limited memory.

In practice, hence, we partition the training corpus and define a subset $V'$ of
the target vocabulary for each partition prior to training. Before training
begins, we sequentially examine each target sentence in the training corpus and
accumulate unique target words until the number of unique target words reaches
the predefined threshold $\tau$. The accumulated vocabulary will be used for
this partition of the corpus during training. We repeat this until the end of
the training set is reached. Let us refer to the subset of target words used for
the $i$-th partition by $V'_i$.

This may be understood as having a separate proposal distribution $Q_i$ for each
partition of the training corpus. The distribution $Q_i$ assigns equal
probability mass to all the target words included in the subset $V'_i$, and zero
probability mass to all the other words, i.e.,
\begin{align*}
    Q_i(y_k) = \begin{cases}
        \frac{1}{\left| V'_i \right|} & \mbox{ if } y_t \in V'_i \\
        0 & \mbox{ otherwise}.
    \end{cases}
\end{align*}
This choice of proposal distribution cancels out the correction term $- \log
Q(y_k)$ from the importance weight in
Eqs.~\eqref{eq:neg_grad_approx}--\eqref{eq:importance_weight}, which makes the
proposed approach equivalent to approximating the exact output probability in
Eq.~\eqref{eq:real_output} with
\begin{align*}
    \label{eq:real_output}
    p(y_t & \mid y_{<t}, x)  \\
    =& \frac{\exp\left\{ \vw_t^\top \phi\left( y_{t-1},
    z_{t}, c_t\right) + b_t \right\}}
    {
        \sum_{k: y_k \in V'} \exp\left\{ \vw_k^\top \phi\left( y_{t-1},
    z_{t}, c_t\right) + b_k \right\}
    }.
\end{align*}
It should be noted that this choice of $Q$ makes the estimator biased. 

The proposed procedure results in speed up against usual importance sampling,
as it exploits the advantage of modern computers in doing matrix-matrix vs
matrix-vector multiplications.

\subsubsection{Informal Discussion on Consequence}

The parametrization of the output probability in Eq.~\eqref{eq:real_output}  can
be understood as arranging the vectors associated with the target words such
that the dot product between the most likely, or correct, target word's vector
and the current hidden state is maximized. The exponentiation followed by
normalization is simply a process in which the dot products are converted into
proper probabilities. 

As learning continues, therefore, the vectors of all the likely target words
tend to align with each other but not with the others.  This is achieved
exactly by moving the vector of the correct word in the direction of
$\phi\left( y_{t-1}, z_{t}, c_t\right)$, while pushing all the other vectors
away, which happens when the gradient of the logarithm of the exact output
probability in Eq.~\eqref{eq:real_output} is maximized.  Our approximate
approach, instead, moves the word vectors of the correct words and of only a
subset of sampled target words (those included in $V'$).



\subsection{Decoding}
\label{sec:decoding}

Once the model is trained using the proposed approximation, we can use the full
target vocabulary when decoding a translation given a new source sentence.
Although this is advantageous as it allows the trained model to utilize the
whole vocabulary when generating a translation, doing so may be too
computationally expensive, e.g., for real-time applications. 

Since training puts the target word vectors in the space so that they align well
with the hidden state of the decoder only when they are likely to be a correct
word, we can use only a subset of candidate target words during decoding. This
is similar to what we do during training, except that at test time, we do
not have access to a set of correct target words. 

The most na\"ive way to select a subset of candidate target words is to take
only the top-$K$ most frequent target words, where $K$ can be adjusted to meet
the computational requirement. This, however, effectively cancels out the whole
purpose of training a model with a very large target vocabulary. Instead, we can
use an existing word alignment model to align the source and target words in the
training corpus and build a dictionary. With the dictionary, for each source
sentence, we construct a target word set consisting of the $K$-most frequent
words (according to the estimated unigram probability) and, using the
dictionary, at most $K'$ likely target words for each source word. $K$ and $K'$
may be chosen either to meet the computational requirement or to maximize the
translation performance on the development set. We call a subset constructed in
either of these ways a {\it candidate list}.

\subsection{Source Words for Unknown Words}
\label{sec:trans2}

In the experiments, we evaluate the proposed approach with the neural machine
translation model called RNNsearch~\cite{Bahdanau-et-al-arxiv2014} (see
Sec.~\ref{sec:rnnsearch}). In this model, as a part of decoding process, we
obtain the alignments between the target words and source locations via the
alignment model in Eq.~\eqref{eq:alignment_weight}. 

We can use this feature to infer the source word to which each target word was
most aligned (indicated by the largest $\alpha_t$ in
Eq.~\eqref{eq:alignment_weight}). This is especially useful when the model
generated an $\left[ \mbox{UNK} \right]$ token. Once a translation is generated
given a source sentence, each $\left[ \mbox{UNK} \right]$ may be replaced using
a translation-specific technique based on the aligned source word. For instance,
in the experiment, we try replacing each $\left[ \mbox{UNK} \right]$
token with the aligned source word or its most likely translation determined by
another word alignment model. Other techniques such as transliteration may also
be used to further improve the performance~\cite{Koehn2010}.

\section{Experiments}

We evaluate the proposed approach in English$\to$French and English$\to$German
translation tasks. We trained the neural machine translation models using only
the bilingual, parallel corpora made available as a part of WMT'14. For each
pair, the datasets we used are:
\begin{itemize}
    \setlength\itemsep{0em}
    \item English$\to$French\footnote{The preprocessed data can be found and
            downloaded from
            \url{http://www-lium.univ-lemans.fr/~schwenk/nnmt-shared-task/README}.}: \\
            {\small \mbox{Europarl v7}, \mbox{Common Crawl}, \mbox{UN}, \\ \mbox{News Commentary}, Gigaword}
    \item English$\to$German: \\
        {\small Europarl v7, Common Crawl, News Commentary}
\end{itemize}

To ensure fair comparison, the English$\to$French corpus, which comprises
approximately 12 million sentences, is identical to the one used in
\cite{Kalchbrenner2013,Bahdanau-et-al-arxiv2014,Sutskever-et-al-NIPS2014}.  As
for English$\to$German, the corpus was preprocessed, in a manner similar to
\cite{peitz-EtAl:2014:W14-33,li-EtAl:2014:W14-331}, in order to remove many
poorly translated sentences.

We evaluate the models on the WMT'14 test set (news-test 2014)\footnote{To
compare with previous submissions, we use the filtered test sets.}, while the
concatenation of news-test-2012 and news-test-2013 is used for model selection
(development set). Table~\ref{tbl:coverage} presents data coverage w.r.t. the
vocabulary size, on the target side.

Unless mentioned otherwise, all reported BLEU scores
\cite{Papineni:2002:BMA:1073083.1073135} are computed with the multi-bleu.perl
script\footnote{\url{https://github.com/moses-smt/mosesdecoder/blob/master/scripts/generic/multi-bleu.perl}}
on the cased tokenized translations.

\begin{table}
    \centering
    \begin{tabular}{l || c | c || c | c }
        & \multicolumn{2}{c||}{English-French} & \multicolumn{2}{c}{English-German}\\
        \hline
        \hline
        & Train & Test & Train & Test\\
        \hline
        \hline
        15k & 93.5 & 90.8 & 88.5 & 83.8\\
        \hline
        30k & 96.0 & 94.6 & 91.8 & 87.9\\
        \hline
        50k & 97.3 & 96.3 & 93.7 & 90.4\\
        \hline
        500k & 99.5 & 99.3 & 98.4 & 96.1\\
        \hline
        All & 100.0 & 99.6 & 100.0 & 97.3\\
    \end{tabular}
    \caption{Data coverage (in \%) on target-side corpora for different
    vocabulary sizes. "All" refers to all the tokens in the training set.}
    \label{tbl:coverage}
    \vspace{-4mm}
\end{table}

\subsection{Settings}

As a baseline for English$\to$French translation, we use the {\bf RNNsearch}
model proposed by \cite{Bahdanau-et-al-arxiv2014}, with 30,000 source and target
words\footnote{The authors of \cite{Bahdanau-et-al-arxiv2014} gave us access to their trained models. We chose the best one on the validation set and resumed training.}. Another RNNsearch model is trained for English$\to$German translation with
50,000 source and target words.

For each language pair, we train another set of RNNsearch models with much
larger vocabularies of 500,000 source and target words, using the proposed
approach. We call these models {\bf \mbox{RNNsearch-LV}}. We vary the size of
the shortlist used during training ($\tau$ in Sec.~\ref{sec:algorithm}). We
tried 15,000 and 30,000 for English$\to$French, and 15,000 and 50,000 for
English$\to$German. We later report the results for the best performance on the development set, with models generally evaluated every twelve hours.

For both language pairs, we also trained new models, with $\tau=15,000$ and
$\tau=50,000$, by reshuffling the dataset at the beginning of each epoch.
While this causes a non-negligible amount of overhead, such a change allows
words to be contrasted with different sets of other words each epoch.

To stabilize parameters other than the word embeddings, at the end of the training
stage, we freeze the word embeddings and tune only the other parameters for
approximately two more days after the peak performance on the development set is
observed. 
This 
helped increase BLEU scores on the development set.

We use beam search to generate a translation given a source. During beam search,
we keep a set of 12 hypotheses and normalize probabilities by the length of the
candidate sentences, as in \cite{Cho2014a}\footnote{These experimental details differ from \cite{Bahdanau-et-al-arxiv2014}.}. The candidate list is chosen to maximize the performance
on the development set, for $K\in\{15k, 30k, 50k\}$ and $K'\in\{10,20\}$.
As explained in Sec.~\ref{sec:decoding}, we test using a bilingual
dictionary to accelerate decoding and to replace unknown words in translations.
The bilingual dictionary is built using \textit{fast\_align} \cite{Dyer2013}.
We use the dictionary only if a word starts with a lowercase letter, and
otherwise, we copy the source word directly. This led to better performance on
the development sets.

\begin{table*}
    \centering
    \begin{tabular}{| l || c | c | c | c | c | c}
        & RNNsearch & RNNsearch-LV & Google & \multicolumn{2}{c|}{Phrase-based SMT} \\
        \hline
        \hline
        Basic NMT & 29.97 {\small (26.58)} & 32.68 {\small (28.76)} & 30.6$^\star$ &
        \multirow{5}{*}{33.3$^\ast$} & \multirow{5}{*}{37.03$^\bullet$} \\
        \cline{1-4}
        +Candidate List & -- & 33.36 {\small (29.32)} & -- & & \\
        +UNK Replace & 33.08 {\small (29.08)}  & 34.11 {\small (29.98)} & 33.1$^\circ$ & & \\
        \cline{1-4}
        +Reshuffle ($\tau$=50k) & -- & 34.60 {\small (30.53)}& --  & & \\
        \cline{1-4}
        +Ensemble & -- & 37.19 {\small (31.98)}& 37.5$^\circ$  & &
    \end{tabular}

    (a) English$\to$French

    \vspace{3mm}

    \begin{tabular}{| l || c | c | c |}
        & RNNsearch & RNNsearch-LV & Phrase-based SMT \\
        \hline
        \hline
        Basic NMT & 16.46 {\small (17.13)} & 16.95 {\small (17.85)} & \multirow{5}{*}{20.67$^\diamond$} \\
        \cline{1-3}
        +Candidate List & -- & 17.46 {\small (18.00)} & \\
        +UNK Replace & 18.97 {\small (19.16)} & 18.89 {\small (19.03)} & \\
        \cline{1-3}
        +Reshuffle & -- & 19.40 {\small (19.37)}& \\
        \cline{1-3}
        +Ensemble & -- & 21.59 {\small (21.06)}& 
    \end{tabular}

    (b) English$\to$German

    \caption{The translation performances in BLEU obtained by different models
        on (a) English$\to$French and (b) English$\to$German translation tasks.
        RNNsearch is the model proposed in \cite{Bahdanau-et-al-arxiv2014},
        RNNsearch-LV is the RNNsearch trained with the approach proposed in this
        paper, and Google is the LSTM-based model proposed in
        \cite{Sutskever-et-al-NIPS2014}. Unless mentioned otherwise, we report
        single-model RNNsearch-LV scores using $\tau=30,000$ (English$\to$French)
        and $\tau=50,000$ (English$\to$German). For the experiments we have run
        ourselves, we show the scores on the development set as well in the
        brackets.
        ($\star$) \cite{Sutskever-et-al-NIPS2014}, ($\circ$) \cite{Luong2014},
        ($\bullet$) \cite{Durrani2014}, ($\ast$) Standard Moses Setting
        \cite{Cho-et-al-EMNLP2014}, ($\diamond$) \cite{Buck2014}.}
    \label{tbl:result_bleu}
\vspace{-4mm}
\end{table*}

\subsection{Translation Performance}

In Table.~\ref{tbl:result_bleu}, we present the results obtained by the trained
models with very large target vocabularies, and alongside them, the previous
results reported in \cite{Sutskever-et-al-NIPS2014}, \cite{Luong2014},
\cite{Buck2014} and \cite{Durrani2014}. Without translation-specific
strategies, we can clearly see that the RNNsearch-LV
outperforms the baseline RNNsearch. 


In the case of the English$\to$French task, RNNsearch-LV approached the
performance level of the previous best single neural machine translation (NMT)
model, even without any translation-specific techniques
(Sec.~\ref{sec:decoding}--\ref{sec:trans2}).  With these, however, the
RNNsearch-LV outperformed it. The performance of the RNNsearch-LV is also
better than that of a standard phrase-based translation
system~\cite{Cho-et-al-EMNLP2014}.  Furthermore, by combining 8 models, we were
able to achieve a translation performance comparable to the state of the art,
measured in BLEU.

For English$\to$German, the RNNsearch-LV outperformed the baseline before unknown
word replacement, but after doing so, the two systems performed similarly. We could
reach higher large-vocabulary single-model performance by reshuffling the dataset,
but this step could potentially also help the baseline. In this case,
we were able to surpass the previously reported best translation result on this task
by building an ensemble of 8 models.

With $\tau=15,000$, the RNNsearch-LV performance worsened a little, with best
BLEU scores, without reshuffling, of 33.76 and 18.59 respectively for
English$\to$French and English$\to$German.

\begin{table}
    \centering
    \begin{tabular}{l || c | c }
        & CPU$^\star$ & GPU$^\circ$ \\
        \hline
        \hline
        RNNsearch & {0.09} s & {0.02} s\\
        \hline
        RNNsearch-LV & {0.80} s & {0.25} s\\
        \hline
        RNNsearch-LV & \multirow{2}{*}{0.12 s} & \multirow{2}{*}{0.05 s}\\
        +Candidate list &  & \\
    \end{tabular}
    \caption{The average per-word decoding time. Decoding here does not include
        parameter loading and unknown word replacement. The baseline uses 30,000 words.
        The candidate list is built with $K$ = 30,000 and $K'$ = 10.\\
        ($\star$) i7-4820K (single thread), 
        ($\circ$) GTX TITAN Black}
    \label{tbl:decoding_speed}
    \vspace{-4mm}
\end{table}

\subsection{Note on Ensembles}


For each language pair, we began training four models from each of which two
points corresponding to the best and second-best performance on the development
set were collected. We continued training from each point, while keeping the word
embeddings fixed, until the best development performance was reached, and took
the model at this point as a single model in an ensemble. This procedure
resulted
in total eight models, but because much of training had been shared, the
composition of the ensemble may be sub-optimal.  This is supported by the fact
that higher cross-model BLEU scores~\cite{freitag2014eu} are observed for models
that were partially trained together.

\subsection{Analysis}

\subsubsection{Decoding Speed}

In Table~\ref{tbl:decoding_speed}, we present the timing information of decoding
for different models. Clearly, decoding from RNNsearch-LV with the full target
vocabulary is slowest. If we use a candidate list for decoding each translation, the
speed of decoding substantially improves and becomes close to the baseline
RNNsearch.

A potential issue with using a candidate list is that for each source sentence,
we must re-build a target vocabulary and subsequently replace a part of the
parameters, which may easily become time-consuming. We can address this issue,
for instance, by building a common candidate list for multiple source sentences.
By doing so, we were able to match the decoding speed of the baseline RNNsearch
model.

\subsubsection{Decoding Target Vocabulary}

\begin{figure}[t]
    \centering
    \begin{minipage}{0.53\textwidth}
        \includegraphics[width=\textwidth,clip=True]{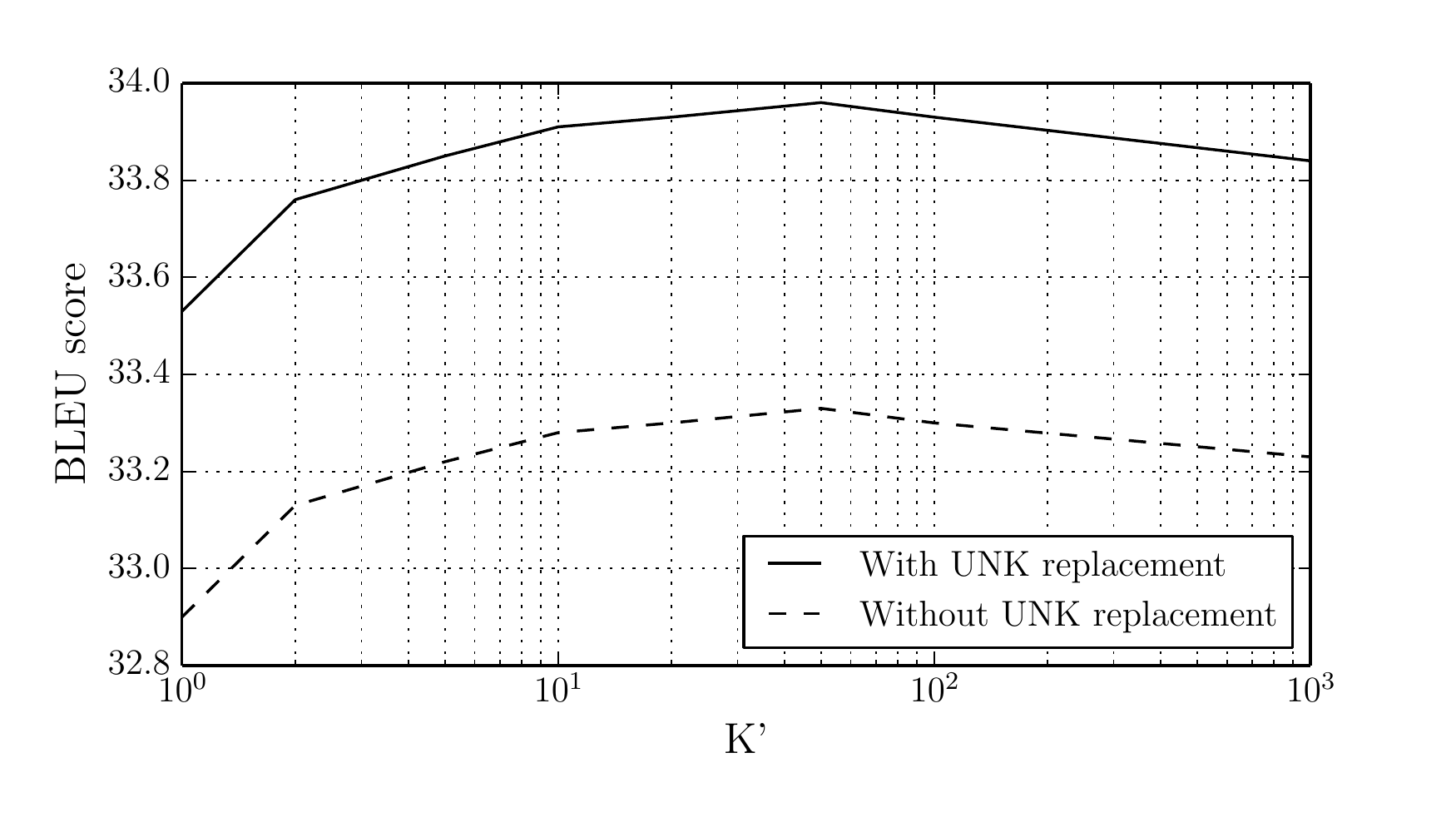}
    \end{minipage}
    \hfill
    \caption{
    Single-model test BLEU scores (English$\to$French) with respect to the number of
    dictionary entries $K'$ allowed for each source word.
    }
    \label{fig:bleu_Kprime}
\end{figure}

For English$\to$French $\left(\tau=30,000\right)$,
we evaluate the influence of the target vocabulary when
translating the test sentences by using the union of a fixed set of $30,000$
common words and (at most) $K'$ likely candidates for each source word
according to the dictionary. Results are presented in Figure
\ref{fig:bleu_Kprime}. With $K'=0$ (not shown), the performance of the system
is comparable to the baseline when not replacing the unknown words (30.12), but
there is not as much improvement when doing so (31.14). As the large vocabulary
model does not predict $\left[ \mbox{UNK} \right]$ as much during training, it
is less likely to generate it when decoding, limiting the effectiveness of the
post-processing step in this case. With $K'=1$, which limits the diversity of
allowed uncommon words, BLEU is not as good as with moderately larger $K'$,
which indicates that our models can, to some degree, correctly choose between
rare alternatives. If we rather use $K=50,000$, as we did for testing based on
validation performance, the improvement over $K'=1$ is approximately 0.2 BLEU.

When validating the choice of $K$, we found it to be correlated to which $\tau$
was chosen during training. For example, on the English$\to$French validation set,
with $\tau=15,000$ (and $K'=10$), the BLEU score is 29.44 with $K=15,000$, but drops
to $29.19$ and $28.84$ respectively for $K=30,000$ and $50,000$. For $\tau=30,000$,
scores increase moderately from $K=15,000$ to $K=50,000$. Similar effects were observed
for English$\to$German and on the test sets. As our implementation of importance
sampling does not apply to usual correction to the gradient, it seems beneficial
for the test vocabularies to resemble those used during training.

\section{Conclusion}

In this paper, we proposed a way to extend the size of the target vocabulary for
neural machine translation. The proposed approach allows us to train a model
with much larger target vocabulary without any substantial increase in
computational complexity. It is based on the earlier work in
\cite{Bengio+Senecal-2008} which used importance sampling to reduce the
complexity of computing the normalization constant of the output word
probability in neural language models. 

On English$\to$French and English$\to$German translation tasks, we observed that
the neural machine translation models trained using the proposed method
performed as well as, or better than, those using only limited sets of target
words, even when replacing unknown words. As performance of the RNNsearch-LV
models increased when only a selected subset of the target vocabulary was used
during decoding, this makes the proposed learning algorithm more practical.

When measured by BLEU, our models showed translation performance comparable to
the state-of-the-art translation systems on both the English$\to$French task and
English$\to$German task. On the English$\to$French task, a model trained with
the proposed approach outperformed the best single neural machine translation
(NMT) model from \cite{Luong2014} by approximately 1 BLEU point.  The
performance of the ensemble of multiple models, despite its relatively less
diverse composition, is approximately 0.3 BLEU points away from the best
system~\cite{Luong2014}.  On the English$\to$German task, the best performance
of 21.59 BLEU by our model is higher than that of the previous state of the art
(20.67) reported in \cite{Buck2014}.

\section*{Acknowledgments}

The authors would like to thank the developers of
Theano~\cite{bergstra+al:2010-scipy,Bastien-Theano-2012}.  We  acknowledge the
support of the following agencies for research funding and computing support:
NSERC, Calcul Qu\'{e}bec, Compute Canada, the Canada Research Chairs and CIFAR.

\bibliography{myref,ml,aigaion}
\bibliographystyle{acl}

\end{document}